\documentclass[letterpaper]{article} 
\usepackage{aaai2026}  
\nocopyright
\usepackage{times}  
\usepackage{helvet}  
\usepackage{courier}  
\usepackage[hyphens]{url}  
\usepackage{graphicx} 
\usepackage{natbib}  
\usepackage{caption} 
\usepackage{algorithm}
\usepackage{algorithmic}
\usepackage{amsmath}
\usepackage{multirow}
\usepackage{colortbl}
\usepackage{booktabs}
\usepackage[table]{xcolor} 
\newcommand{\eg}{e.g.}
\newcommand{\ie}{i.e.}
\usepackage{bm}
\usepackage{amssymb}
\newcolumntype{M}[1]{>{\centering\arraybackslash}p{#1}}

\usepackage{newfloat}
\usepackage{listings}
\DeclareCaptionStyle{ruled}{labelfont=normalfont,labelsep=colon,strut=off} 
\floatstyle{ruled}
\newfloat{listing}{tb}{lst}{}
\floatname{listing}{Listing}
\title{Gene Incremental Learning for Single-Cell Transcriptomics\thanks{This paper has been accepted by AAAI 2026.}}
\author {
    Jiaxin Qi\textsuperscript{\rm 1}, 
    Yan Cui\textsuperscript{\rm 2}, 
    Jianqiang Huang\textsuperscript{\rm 1,\rm 2,\rm 3}\protect\thanks{Corresponding author.}, 
    Gaogang Xie\textsuperscript{\rm 1,\rm 3}
}
\affiliations {
    \textsuperscript{\rm 1}Computer Network Information Center, Chinese Academy of Sciences, Beijing, China\\
    \textsuperscript{\rm 2}Hangzhou Institute for Advanced Study, University of Chinese Academy of Sciences, Hangzhou, China\\ 
    \textsuperscript{\rm 3}University of Chinese Academy of Sciences, Beijing, China
\\
    jxqi@cnic.cn, cuiyan.ch@gmail.com, jqhuang@cnic.cn, xie@cnic.cn
}

\begin{document}

\maketitle


\begin{abstract}
Classes, as fundamental elements of Computer Vision, have been extensively studied within incremental learning frameworks. In contrast, tokens, which play essential roles in many research fields, exhibit similar characteristics of growth, yet investigations into their incremental learning remain significantly scarce. 
This research gap primarily stems from the holistic nature of tokens in language, which imposes significant challenges on the design of incremental learning frameworks for them. 
To overcome this obstacle, in this work, we turn to a type of token, gene, for a large-scale biological dataset---single-cell transcriptomics---to formulate a pipeline for gene incremental learning and establish corresponding evaluations.  
We found that the forgetting problem also exists in gene incremental learning, thus we adapted existing class incremental learning methods to mitigate the forgetting of genes. Through extensive experiments, we demonstrated the soundness of our framework design and evaluations, as well as the effectiveness of our method adaptations. 
Finally, we provide a complete benchmark for gene incremental learning in single-cell transcriptomics.
\end{abstract}

\begin{links}
    \link{Code}{https://github.com/simpleshinobu/scbenchmark}
\end{links}

\section{Introduction}
\label{sec:intro}

\begin{figure}[t!]
\centering\includegraphics[width=\linewidth]{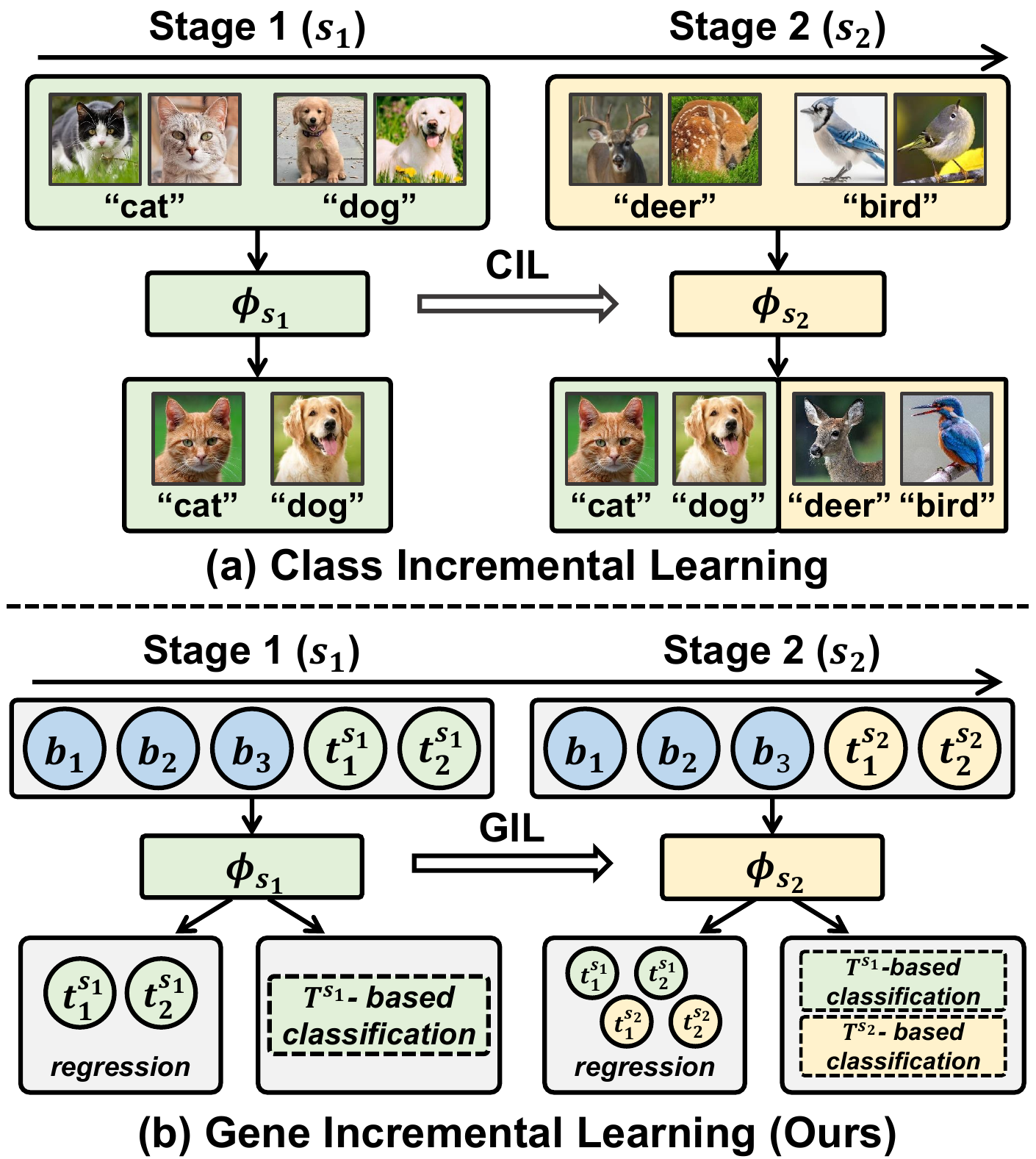}
    \caption{Illustrations of (a) Class Incremental Learning (CIL) framework and (b) our proposed Gene Incremental Learning (GIL) framework. In CIL, the given classes are exclusive at each stage, and classification accuracy is tested across all previously seen classes. In GIL, $b_i, i=1,2,3,\ldots$ denote the base tokens given in every stage, while $T^{s_i}=\{t^{s_i}\}$ represents the set of specific tokens to be learned in stage $i$. For evaluation, regression refers to the token-wise regression loss, and $T^{s_i}$-based classification denotes performing the classification on the specific downstream dataset where the token set $T^{s_i}$ is crucial.} 
    \label{fig:teaser}
    \vspace{-5pt}
\end{figure}

The class of an object serves as a foundational concept in Computer Vision. It is observed that the number of classes often increases due to factors such as the discovery of new species in the natural world and the assignment of novel class labels to recently developed objects. In response, the Class Incremental Learning (CIL) framework has been introduced to evaluate a model's ability to continuously learn new classes~\cite{wu2019large,zhang2020class,masana2022class}.
The pipeline for this framework is illustrated in Figure~\ref{fig:teaser}(a). Initially, the model trains on images from specific classes, \ie, ``cat'' and ``dog''. Subsequent stages of training focus exclusively on datasets containing new classes, such as ``deer'' and ``bird'', ensuring that samples from previous classes remain inaccessible during these stages. Evaluations will be performed across all seen classes until the current stage. Considering the specified framework, extensive studies highlight catastrophic forgetting of previous classes as the crucial challenge in Class Incremental Learning. To mitigate such forgetting, methods designed on data replay~\cite{zhu2021class,hu2021distilling} and knowledge distillation~\cite{dong2021few,kang2022class} have been extensively proposed and explored.

Similarly, tokens, crucial elements for many fields~\cite{pennington2014glove,scGPT,scFoundation}, also exhibit growth characteristics like classes. For example, in natural language processing, where tokens represent words, the continual invention of new words leads to an expansion of the vocabulary~\cite{lehrer2003understanding}. Likewise, in the biological field of single-cell transcriptomics, where tokens represent genes, new genes are continually discovered due to advancements in measurement technologies~\cite{karaayvaz2018unravelling,aljanahi2018introduction}, contributing to the expansion of the gene pool. Thus, incremental learning for tokens has practical significance. However, this framework has been consistently overlooked, primarily due to the challenges in defining it within the holistic nature of language data. For example, if we apply the settings of CIL to divide different words into different stages, \eg, the word ``learning'' no longer appears in one stage, it becomes impractical to either collect texts that do not contain the word ``learning'', which would significantly reduce the amount of data, or just remove the word ``learning'' from existing texts, which would change the original meaning. Moreover, these difficulties become significantly exacerbated when multiple works need to be excluded.

Fortunately, the challenge mentioned earlier does not exist in single-cell transcriptomics (simplified as transcriptomics)~\cite{tang2009mrna}, allowing us to design the gene incremental learning framework to address the increase of new genes.
In transcriptomics, genes are viewed as tokens, similar to words, and each sample consists of a sequence of gene expression values, analogous to a sentence, and the mainstream models in this field are based on Transformers~\cite{scGPT}.
Unlike the holistic nature of language, transcriptomic data lacks relative orders among genes, allowing for straightforward division and rearrangement of genes in different incremental stages to establish an incremental framework. Therefore, we design the Gene Incremental Learning (GIL) framework for transcriptomics with the following details: As shown in Figure~\ref{fig:teaser}(b), we maintain some genes as base genes, which is essential for rendering samples meaningful under transcriptomic contexts. Then, we divide the remaining genes into various stages, ensuring they are mutually exclusive across stages. For example, in stage one, the samples contain base genes and genes $t^{s_1}$, while in stage two, models can only see base genes and genes $t^{s_2}$. Evaluations will be performed across all seen genes until the current stage. This framework effectively constructs a pipeline for Gene Incremental Learning and enables the assessment of the model’s ability to continually learn new genes.

In addition, we propose comprehensive evaluations for Gene Incremental Learning. First, we introduce a gene-wise regression metric that directly assesses model forgetting for previous genes. Second, as shown in Figure~\ref{fig:teaser}(b), we propose a gene-based classification evaluation, where specific genes are selected for each stage, whose learning is crucial for the corresponding downstream classification datasets. This means learning for such genes in a stage will make the model perform better in the downstream classifications associated with that stage. Utilizing these datasets allows us to use classification accuracy to demonstrate whether genes have been memorized or forgotten. Furthermore, to mitigate gene forgetting, we have adopted several fundamental CIL methods to establish the baseline methods for GIL. Through extensive experiments, we validate the rationality of our Gene Incremental Learning framework, the consistency of our evaluation methods, and the effectiveness of our adapted methods. Ultimately, we present a straightforward yet comprehensive benchmark for Gene Incremental Learning for single-cell transcriptomics.

We summarize our main contributions as the following three aspects:
\begin{enumerate}
    \item We thoroughly define the Gene Incremental Learning framework, using single-cell transcriptomic datasets, which addresses the research gap in incremental learning in the context of the continuous growth of genes.

    \item We propose the evaluations for Gene Incremental Learning by introducing a gene-wise regression and gene-based classification to facilitate a thorough assessment of gene learning and forgetting within the GIL framework.
    
    \item We adapt existing Class Incremental Learning methods to the GIL and validate the effectiveness of the adaptations through extensive experiments. Finally, we introduce a comprehensive Gene Incremental Learning benchmark for single-cell transcriptomics.
\end{enumerate}

\section{Related Works}
\label{sec:relatedworks}
\subsection{Class Incremental Learning}

Class Incremental Learning (CIL) \cite{chen2018lifelong,pentina2016theoretical}
, also known as lifelong learning~\cite{silver2002task,silver2013lifelong} or continual learning~\cite{shi2024continual,de2021continual}, is inspired by the continual learning pattern observed in human brains~\cite{constantinescu2016organizing,mccaffary2021towards}. It involves training models sequentially on a series of classes while maintaining overall performance on all seen classes. Researchers have identified catastrophic forgetting as the major challenge for CIL~\cite{goodfellow2013empirical,mccloskey1989catastrophic}. To address this issue, two main camps of methods have been proposed: 1). Data replay~\cite{castro2018end,hou2019learning,zhao2020maintaining} demonstrates strong resistance to forgetting by storing exemplars of old classes. 2). Knowledge distillation~\cite{hinton2015distilling,rebuffi2017icarl} retains model behavior by learning the outputs or features of the old models.

In addition to classic CIL, incremental learning also encompasses Task Incremental Learning~\cite{qin2021lfpt5,ke2022continual} and Domain Incremental Learning~\cite{lu2018learning}, which divide tasks or different distributions of data into different incremental stages, respectively. However, these incremental learning frameworks are mainly focused on the increase of class or data, but do not discuss tokens. In this work, we follow the CIL settings, leveraging single-cell transcriptomic data, to define the Gene Incremental Learning, which is fundamentally different from the traditional incremental frameworks.

\subsection{Single-Cell Transcriptomics}

Single-cell transcriptomics, also known as single-cell RNA sequencing (\textit{i.e.}, scRNA-seq), was initially developed by the Surani Lab~\cite{tang2009mrna}. Additionally, the landscape of computational tools and public data repositories for scRNA-seq has rapidly expanded~\cite{voigt2021single,kharchenko2021triumphs}. Today, scRNA-seq is extensively utilized in human health research, primarily to characterize cell types across various organs~\cite{ramachandran2020single,gustafsson2022role} or clarify temporal processes such as human tissue development~\cite{olaniru2023single,collin2021single}.

As genes can be viewed as tokens, researchers have applied NLP methods to transcriptomics, particularly using Transformers as feature extractors. scBERT~\cite{scBERT} was a pioneer in proposing a single-cell pre-training framework utilizing Transformers. Subsequent studies have focused on increasing the data volume~\cite{scGPT}, expanding dataset diversity~\cite{genecompass2023}, and modifying Transformer architectures~\cite{scFoundation,geneFormer2023}. However, researchers have overlooked the potential of transcriptomics to pioneer token learning. Leveraging the properties of the transcriptomic dataset, we have successfully divided tokens and defined the Token Incremental Learning framework.

\section{Method}
\label{sec:method}

\subsection{Gene Learning in Transcriptomics}

In single-cell transcriptomics, given the sample \((\bm{x}, \bm{v})\! =\! (t_1, t_2, \ldots, t_l; v_1, v_2, \ldots, v_l)\), where $\bm{x}$ denotes genes and $\bm{v}$ denotes corresponding expression values, the gene learning strategy can be realized by masked value prediction~\cite{scBERT,genecompass2023} under a self-supervised framework. Define the training set as \(\mathcal{D}\!=\!\{\bm{x}_i,\bm{v}_i\}_{i=1}^N\), and the output is the predictions for the masked values. Then, the loss function can be written as:
\begin{align}
\label{eq:tran_objective}
\mathcal{L}_{\text{tran}}(\mathcal{D},\phi) = \frac{1}{N} \sum_{i=1}^N \sum_{j} \left\|v_{ij} - \hat{v}_{ij} \right\|^2,
\end{align}
where $\hat{v}_{ij}$ represents the corresponding predicted values, $v_{ij}$ denotes the ground-truth value for $j$-th masked value in $\tilde{\bm{v}}_i$ and $\tilde{\bm{v}}_i$ denotes the masked input values. We will further elaborate on the value prediction process by Eq.~\eqref{eq:embed_process} to Eq.~\eqref{eq:value_prediction}, to demonstrate that learning the values associated with genes is equivalent to learning the genes themselves.

\subsection{Gene Incremental Learning Formulation}

\begin{figure*}[t!]
\centering\includegraphics[width=6.4in]{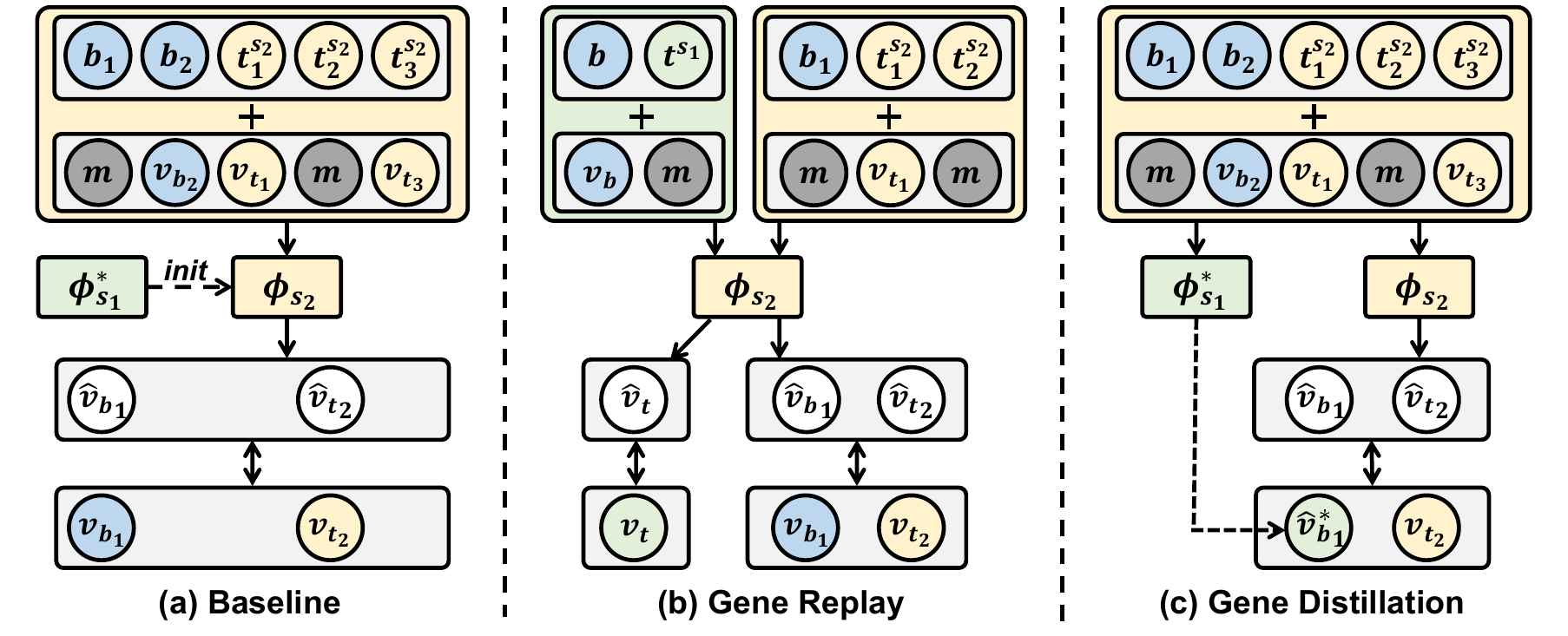}
    \caption{Illustrations of baseline methods for GIL in stage $k$, where we use $k=2$ as an example, and the samples from stage 2 are in yellow background. (a) The baseline shows the masked token prediction loss formulated in Eq.~\eqref{eq:tran_objective}. \textit{init} denotes that the current model $\phi_{s_2}$ is initialized by the previous optimal model $\phi^*_{s_1}$ (b) Data Replay shows that some previous samples (with a green background) are maintained for training in the current stage. (c) Token Distillation shows how the previous optimal model distills knowledge through base token regression, which is formulated in Eq.~\eqref{eq:distillation}.} 
    \label{fig:method}
    \vspace{-8pt}
\end{figure*}

\noindent\textbf{Revisit Class Incremental Learning (CIL).} CIL is designed to enable the models to progressively learn new classes, naturally introducing the concept of stages $S\!=\!(s_1,s_2, \ldots, s_n)$. In each stage, the model should learn different classes and thus CIL separates all classes $Y$ into different stages $Y=(Y^{s_1},Y^{s_2}, \ldots,Y^{s_n})$, where $Y=Y^{s_1} \cup Y^{s_2} \cup \cdots \cup Y^{s_n}$ and the classes designated to different stages are disjoint, \ie, $Y^{s_i} \cap Y^{s_j} = \varnothing, i\neq j, i,j = 1,2,\ldots,n$. Since each class corresponds to specific samples in the classification dataset $\mathcal{D}$, distributing classes across various stages effectively partitions the dataset $\mathcal{D}=(\mathcal{D}^{s_1},\mathcal{D}^{s_2},\ldots,\mathcal{D}^{s_n})$, where $\mathcal{D}^{s_k}\!=\!\{(x,y),y \in Y^{s_k}\}, k\!=\!1,2,\ldots,n$. 

The ultimate goal of CIL is to continually build a classification model for all seen classes. In other words, the model should not only learn classes from the current datasets but also preserve the classification ability learned from former datasets. Formally the objective function for the model $\phi$ in stage $k$ is usually written as:
\begin{align}
\label{eq:cil}
\mathcal{L}_{\text{CIL},s_k}=\mathcal{L}(\mathcal{D}^{s_k},\phi)+\mathcal{L}(\textstyle\bigcup_{i=1}^{k-1}\mathcal{D}^{s_i},\phi),
\end{align}
where \( \mathcal{L}(\mathcal{D}^{s_k}, \phi) \) represents the loss for the current dataset, designed to evaluate the classification ability of classes \( Y^{s_k} \). The term \( \mathcal{L}(\textstyle\bigcup_{i=1}^{k-1}\mathcal{D}^{s_i}, \phi) \) quantifies the risk of model \( \phi \) when performing previous datasets. Due to the invisibility of previous data, this term cannot be directly computed. Thus, it is typically implemented as an approximate constraint, such as mimicking the optimal model in the last stage $\phi^{*}_{s_{k-1}}$, which approximately reflects the former data distributions. 

A good CIL model that minimizes Eq.~\eqref{eq:cil} for every stage finally demonstrates discriminability across all classes, thereby fulfilling the initial goal for CIL.\\

\noindent\textbf{Gene Incremental Learning (GIL).} Inspired by the formulation of CIL, we define the Gene Incremental Learning framework, where our motivation is to enable the model to continuously learn genes in the context of single-cell transcriptomics. Although the transcriptomic data avoids the holistic nature of language when designing the incremental framework, there remain two significant challenges compared to CIL. First, unlike the direct correspondence between classes and samples, genes and samples do not align; each sample consists of all genes and their expression values, and thus dividing genes into different stages does not automatically partition datasets. Second, there is a significant difference in the roles of genes and classes; a class can represent the meaning of a sample as a standalone unit, while a gene and its value, being a single component of the sample, lacks the ability to express the sample's overall meaning.

To address the above challenges, we propose the separate partitioning of datasets and genes, along with a base gene mechanism. To be specific, given a dataset $\mathcal{D}$ and a gene set $T$, we partition both of them into different stages: $\mathcal{D} \!=\! (\mathcal{D}^{s_1}, \mathcal{D}^{s_2}, \ldots, \mathcal{D}^{s_n})$ and $T = ((B, T^{s_1}), (B, T^{s_2}), \ldots, (B, T^{s_n}))$. Here, some genes are designated as base genes $B$, which consistently appear in each stage, while the remaining genes are specifically assigned to different stages such that $T^{s_i} \cap T^{s_j}\! = \!\varnothing$, $i \neq j$ and $T = B \cup T^{s_1} \cup T^{s_2} \cup \cdots \cup T^{s_n}$. The base gene mechanism enables valid gene learning in each stage, reducing the risk of constructing meaningless samples with insufficient genes. For stage $k$, as shown in Figure~\ref{fig:teaser}, the dataset is represented as $\mathcal{D}^{s_k}\!=\{\!(b_1, b_2, \ldots, t_1,t_2,\ldots)_j\}, b_i\in B, t_{i}\in T^{s_k}$. In transcriptomics, values and genes correspond one-to-one. Thus, when genes are determined in each stage, the associated values are correspondingly divided, thereby we omit the notation of values here.
Following Eq.~\eqref{eq:cil}, the GIL objective function for model $\phi$ in stage $k$ is defined as:
\begin{align}
\label{eq:til}
\mathcal{L}_{\text{GIL},s_k} = & \ \mathcal{L}(\mathcal{D}^{s_k},T^{s_k},\phi) +\sum_{i=1}^{k-1}\mathcal{L}(\mathcal{D}^{s_i},T^{s_i},\phi),
\end{align}
where \( \mathcal{L}(\mathcal{D}^{s_k}, T^{s_k}, \phi) \) represents the loss associated with the current dataset, which can be implemented by Eq.~\eqref{eq:tran_objective}. The second term also cannot be explicitly calculated and it is estimated to ensure performance across all seen genes. Note that both gene and dataset partitions could be randomized; alternatively, the partitioning of genes can follow a specific order. For example, we deliberately selected genes, which are crucial for downstream datasets, and assigned them to different stages to align the downstream datasets with stages for our proposed evaluation.

\subsection{Incremental Learning Method Adaptations}

To establish a comprehensive benchmark for our proposed Gene Incremental Learning, we draw inspiration from the methodologies of CIL, adapting several baseline methods to provide foundational directions for this field. For better formulation, we first provide a detailed explanation of feature extraction for genes in transcriptomics. Assume that the current stage is $k$, for a transcriptomic dataset \( \mathcal{D}^{s_k} \), the input is $(\bm{x}, \bm{v})$ and the gene feature extraction process could be formulated as:
\begin{gather}
\label{eq:embed_process}
\bm{e} = \mathbf{E}_{\phi}(\bm{x})  + \tilde{\bm{v}}\mathbf{L}_{1,\phi}, \\
\label{eq:feature_extraction}
\bm{e}' = \mathbf{M}_\phi(\bm{e}), \\
\label{eq:value_prediction}
\hat{\bm{v}} = \bm{e}'\mathbf{L}_{2,\phi},
\end{gather}
where $\tilde{\bm{v}}$ is masked values $\bm{v}$, $\hat{\bm{v}}$ is the predicted values, $\mathbf{E}_{\phi}$ is gene embeddings layer,
$\mathbf{L}_{1,\phi}\!\in\! \mathbb{R}^{1\times d}$ is a linear layer that encodes the values into embeddings, $\mathbf{L}_{2,\phi}\in \mathbb{R}^{d\times 1}$ is a linear layer to predict masked values from encoded features, $d$ is the hidden dimension, $\mathbf{M}_\phi$ is the backbone, which is usually implemented as Transformers. According to Eq.~\eqref{eq:embed_process}, genes are bound to corresponding values, demonstrating learning values are indeed learning genes as we mentioned.\\

\noindent\textbf{Baseline and Oracle.} As shown in Figure~\ref{fig:method}(a), our baseline is defined as optimizing the model only using the current dataset $\mathcal{D}^{s_k}$ for learning $T^{s_k}$ at stage $k$, while ignoring the second term in Eq.~\eqref{eq:til}. The loss can be written as:
\begin{align}
\label{eq:baseline}
\mathcal{L}_{\text{base},s_k} = \mathcal{L}_{\text{tran}}(\mathcal{D}^{s_k}, \phi).
\end{align}
Here we omit the input $T^{s_k}$ as formulated in Eq.~\eqref{eq:til} because $T^{s_k}$ is already bound to $\mathcal{D}^{s_k}$ based on our GIL design. Note that the parameters $\phi$ at each stage are initialized with the optimal parameters trained from the previous stage. Unless otherwise specified, we assume $k>1$ because stage one can only take the baseline training.

To establish an upper bound for reference, we train the model on all datasets ${\{\mathcal{D}^{s_i}\}_{i=1}^{n}}$ derived from GIL splits, as the oracle method:
\begin{align}
\label{eq:oracle}
\mathcal{L}_{\text{oracle}} = \sum_{i=1}^{n} \mathcal{L}_{\text{tran}}(\mathcal{D}^{s_i}, \phi).
\end{align}
The oracle is expected to provide the best global performance across all genes, while for specific genes at a given stage, the performance may not exceed that of the baseline.\\

\noindent\textbf{Gene Replay.} A mainstream method in CIL leverages the incremental settings by retaining a subset of samples from previous datasets for current training, referred as data replay, and in our GIL framework, the gene replay strategy could also be implemented as data replay.
Some advanced methods are proposed such as dataset condensation~\cite{mitra2000data} to further improve the performance. To provide a basic reference for our GIL benchmark, we evaluate the native implementation:
\begin{align}
\label{eq:data_replay}
\mathcal{L}_{\text{dr},s_k} = \mathcal{L}_{\text{tran}}(\mathcal{D}^{s_k}, \phi) + \sum_{i=1}^{k-1}\mathcal{L}_{\text{tran}}(\mathcal{D}_{\text{dr}}^{s_i}, \phi),
\end{align}
where $\mathcal{D}_{\text{dr}}^{s_i} \subset \mathcal{D}^{s_i}$ is a subset for previous dataset $\mathcal{D}^{s_i}$, and the training pipeline in stage $k$ is shown in Figure~\ref{fig:method}(b).\\

\noindent\textbf{Gene Distillation.} Another method in CIL involves distilling knowledge from the previous model, which assumes old models can represent the current sample by old classes. In GIL, we adapt the class distillation to gene distillation, as shown in Figure~\ref{fig:method}(c). According to Eq.~\eqref{eq:tran_objective}, we have: 
\begin{equation}
\label{eq:distillation}
\mathcal{L}_{\text{fd},s_k}\! =\! \frac{1}{N_k} \sum_{i=1}^{N_k}( \sum_{j} {\|v_{ij}} \!-\! \hat{v}_{ij}\|^2 \!+\! \lambda \| \hat{\bm{v}}_{i}\!-\! \hat{\bm{v}}^{*}_{i,s_{k-1}} \|^2),
\end{equation}
where $N_k$ denotes the number of samples in this stage, $\hat{v}_{ij}$ is the prediction for the masked values, $\hat{\bm{v}}^{*}_{i,s_{k-1}}$ denotes the output derived from the optimal model $\phi^*_{s_{k-1}}$ in the last stage, $\lambda$ is the coefficient for distillation. Note that the specific genes for the current stage are removed from the second term in the implementation due to $\phi^*_{s_{k-1}}$ do not have the ability to predict the unseen genes.

\subsection{Evaluations}
\noindent\textbf{Gene-wise Regression.} The most straightforward method to evaluate gene learning is using masked gene prediction loss in Eq.~\eqref{eq:tran_objective}. As multiple genes are learned in a single stage, we average the performance across all genes learned specifically in that stage. For stage $k$, we have:
\begin{align}
\label{eq:Regression}
\mathcal{L}_{\text{regress},s_k} = \mathbb{E}[\sum_k \| v_{ik} - \hat{v}^*_{ik} \|^2], t_{ik} \in T^{s_k},
\end{align}
where $\hat{v}^*_{ik}$ is the predicted masked values by the optimal model trained in the current stage $\phi^*_{s_k}$, $v_{ik}$ is the expression value of its corresponding gene $t_{ik}$, and $T^{s_k}$ denotes the set of learned specific genes in stage $k$.\\

{\scriptsize
\begin{table*}[t]
\centering
\renewcommand{\arraystretch}{0.8}
{
\begin{tabular}{c|c|ccc|ccc|ccc|c}
\toprule
Method & Stage & Norman & Lupus & $\Delta$ & ICol & Adamson & $\Delta$ & Lupus & Panc & $\Delta$ & Avg \\
\midrule
\multirow{2}{*}{Baseline} &  1 
& 0.172 & - & - & 0.164 & - & - & 0.145 & - & -  & - \\
&  2 
& 0.424 & 0.134 & 0.253 & 0.496 & 0.137 & 0.333 & 0.397 & 0.204 & 0.252 & 0.279\\
\midrule
Oracle   & - &  0.173 & 0.136 & - & 0.164 & 0.115 & - & 0.145 & 0.206 & -  & - \\
\midrule
\multirow{2}{*}{Replay}  &  1 & 0.172 & - & -
& 0.164 & - & -   
& 0.146 & - & - & -  \\
&  2 & 0.215 & 0.134 & 0.043 
& 0.200 & 0.124 &  0.036 
& 0.177 & 0.213 & 0.031 &  0.037  \\
\midrule
\multirow{2}{*}{Distill}   &  1 
& 0.172  &-  & -  
& 0.163 & - &  - 
& 0.147 & - &  - & - \\
&  2 & 0.365 & 0.139 & 0.193
& 0.420  & 0.145 &   0.257
&  0.332  & 0.220 &  0.185 & 0.212 \\
\bottomrule
\end{tabular}
}
\vspace{-4pt}
\caption{Averaged regression loss for specific genes in each stage for three 2-stage GIL settings on evaluation set (Gene-wise Regression). The three settings are Norman-Lupus, ICol-Adamson, and Lupus-Panc. The model learns crucial genes ($T^{s_k}$) for the associated dataset at each stage, thus using the name of the dataset to represent corresponding genes. $\Delta$ represents the forgetting of genes learned in the previous stage (here is stage 1), as reflected by the difference in regression loss. Avg denotes the averaged $\Delta$ across three GIL settings. The smaller the absolute value of $\Delta$, the better, and lower regression losses for others are preferred. The default replay number of samples is 1,000 and the default $\lambda$ of distillation is 5.0. ``-'' denotes the result is not applicable. Results are the mean of three independent trials.}
\label{table1}
\end{table*}
}

\noindent\textbf{Gene-based Classification.} 
The above evaluation might not provide a universally comparable measure due to variations in different datasets and value scales. Therefore we design gene-based classification as another evaluation for GIL. Specifically, we identified some crucial genes for different downstream transcriptomics classification tasks and divided these genes into different stages. Then, at each stage, we can assess the gene performance through the corresponding downstream task. For example, for a downstream dataset \(\mathcal{D}_{d_1}\), where \(T^{s_1}\) is crucial for its classification, we learn \(T^{s_1}\) at stage one and then measure how well the model retains \(T^{s_1}\) through tests on \(\mathcal{D}_{d_1}\) in the following stages. The downstream classification loss is written as:
\begin{align}
\label{eq:classification}
\mathcal{L}_{\text{class},s_k} = \frac{1}{N_{d_k}} \sum_{i=1}^{N_{d_k}} - \bm{y}_i\cdot\log p(\bm{e}'^*_{s_k}\mathbf{L}),
\end{align}
where $N_{d_k}$ is the number of samples in downstream dataset $\mathcal{D}_{d_k}$, $\bm{y}_i$ is the one-hot class label, $\mathbf{L}$ is a trainable linear layer to project the feature into class space and $\bm{e}'^*_{s_k}$ is the extracted feature by the optimal model $\phi^*_{s_k}$, formulated in Eq.~\eqref{eq:embed_process} and Eq.~\eqref{eq:feature_extraction}, which is frozen to only extract gene features.

\section{Experiment}
\label{sec:exp}

\subsection{Dataset}

In this paper, we leveraged the data collection method outlined by scGPT~\cite{scGPT}, drawing from the CELLxGENE collection~\cite{megill2021cellxgene,czi2023cz}, which consists of human cell data characterized by gene-expression pairs. This extensive dataset covers over 50 organs and tissues such as blood and heart, derived from more than 400 studies, providing a comprehensive view of cellular diversity within the human body. We randomly selected 906,890 samples for training and 204,871 samples for gene-wise evaluation. We also followed scGPT~\cite{scGPT} to construct the gene vocabulary consisting of 60,697 genes. 

For gene-based downstream classification, we collected six transcriptomic datasets for comprehensive evaluations: {Norman}~\cite{norman2019exploring} explores the relationship between the set of genes expressed by a cell and its phenotype; {Lupus}~\cite{perez2022single} shows an increase in type 1 interferon-stimulated genes; Inhibitor Colitis (ICol)~\cite{thomas2024single} reveals the interactions between circulating T cells and epithelial cells; {Adamson}~\cite{adamson2016multiplexed} applies Perturb-seq to dissect the mammalian unfolded protein response; Pancreas (Panc)~\cite{chen2023transformerPanc} consolidates data from five human pancreas studies; and Myeloid (Myel) ~\cite{cheng2021panMyel} provides a comprehensive pan-cancer analysis of myeloid cells.

{\scriptsize
\begin{table}[!t]
\centering
\renewcommand{\arraystretch}{0.8}
{
\begin{tabular}{c|c|ccc|c}
\toprule
Method & Stage & ICol & Myel & Panc & $\Delta$   \\
\midrule
\multirow{3}{*}{Baseline} &  1 & 0.163 &- &- & -\\
         &  2 & 0.452 & 0.263 &- & 0.289 \\
         &  3 & 0.498 & 0.290 & 0.192 & 0.181 \\
\midrule
Oracle   & - & 0.163 & 0.209 & 0.175 & - \\
\midrule
\multirow{3}{*}{Replay}   &  1 & 0.163 & - & - & - \\
         &  2 & 0.205 & 0.263  & - & 0.042 \\
         &  3 & 0.209 & 0.272 & 0.190 & 0.028 \\
\midrule
\multirow{3}{*}{Distill}  &  1 & 0.164 & - & - & -  \\
         &  2 & 0.437 & 0.248 & - & 0.273 \\
         &  3 & 0.444 & 0.323 & 0.214 & 0.177 \\
\bottomrule
\end{tabular}
}
\vspace{-4pt}
\caption{Averaged regression loss for specific genes at each stage in a 3-stage GIL setting (ICol-Myel-Panc) on the evaluation set. Selected genes for each dataset correspond to the specific genes at each respective stage. $\Delta$ denotes the averaged forgetting for the previous genes associated with their datasets, \eg, $\Delta$ in Stage 3 is calculated by (($\text{ICol}_{s_3}\! -\! \text{ICol}_{s_1})\! +\! (\text{Myel}_{s_3}\! -\! \text{Myel}_{s_2}))/2$, where the subscript denotes the performance of the dataset at that stage. Other settings are the same as those in Table~\ref{table1}.}
\vspace{-6pt}
\label{table3stage}
\end{table}
}
{\scriptsize
\begin{table*}[t]
\centering
\renewcommand{\arraystretch}{0.8}
{
\begin{tabular}{c|c|ccc|ccc|ccc|c}
\toprule
Method & Stage & Norman & Lupus & $\Delta$ & ICol & Adamson & $\Delta$ & Lupus & Panc & $\Delta$ & Avg \\
\midrule
\multirow{2}{*}{Baseline} &  1 
& 37.734 & 67.313 & - & 59.370 & 42.209 & - & 74.451 & 93.542 & - & -\\
&  2 
& 35.590 & 75.386 & -2.144 & 56.771 & 43.514 & -2.599 & 73.747 & 97.826 & -0.704 & -1.816 \\
\midrule
Oracle   & - & 38.112 & 75.422 & - & 59.201 & 43.782 & - & 74.893 & 97.913 & - & -\\
\midrule
\multirow{2}{*}{Replay}  &  1 & 37.734 & 67.315 & -
& 59.366  & 42.209 & -
& 74.456 & 93.550 &  -  & - \\
&  2 & 36.449 & 75.000 &  -1.285 
&  57.738 & 43.618  &   -1.628
& 74.143 & 97.948 &  -0.313 & -1.075 \\
\midrule
\multirow{2}{*}{Distill}   &  1 &37.902 & 67.577 & -
& 59.367 & 42.209  & -   
& 74.659 & 93.565 &  - & - \\
&  2 & 34.162 & 72.939 & -3.740
& 56.347 &  42.480 & -3.020
& 74.000 & 97.441 &  -0.659 & -2.473 \\
\bottomrule
\end{tabular}
}
\vspace{-4pt}
\caption{Test accuracy (\%) for three 2-stage GIL settings on downstream classification datasets (Gene-based Classification). The specific genes $T^{s_k}$ for each stage are crucial for the associated dataset classification (\eg, in the first settings, the specific genes in stage 1 are important for Norman classification). $\Delta$ represents the forgetting of genes, as reflected by the difference in classification accuracy. The smaller the absolute value of $\Delta$, the better, while higher accuracies for others are preferred. Other settings are the same as those in Table~\ref{table1}.}
\label{table2}
\end{table*}
}

\subsection{Implementation Details}

To ensure consistent and fair comparisons, we configured the same model and training parameters for all experiments. We followed the scGPT~\cite{scGPT} and employed a Transformer as the feature extractor with 6 layers, 8 heads for multi-head attention, and hidden dimensions of 256. The experiments were conducted on an 8-NVIDIA A100 GPU server. In the GIL training, we applied a batch size of 128, and the Adam~\cite{kingma2014adam} optimizer with a learning rate of 0.0005 across 5 epochs for each stage, and here a warm-up strategy is applied in the first 5,000 iterations for all methods. We selected important genes for downstream datasets based on their cumulative gene values calculated across all samples in the corresponding datasets and removed duplicate genes across datasets to prevent confusion.

For gene-wise evaluation, we only considered the seen stage-specific genes in each stage and calculated the average loss across these genes. Following scGPT, the length of the input is limited to 512, and genes are randomly selected. Therefore, the evaluations for each trial may contain different genes for each sample. To mitigate randomness, we construct a large-scale evaluation set, and the experimental results demonstrate that gene-wise regression evaluation is stable. For gene-based classification, the GIL model was frozen to extract features, and only a single linear layer was optimized. All experiments were independently conducted three times, and the average performance was reported. More details can be found in the Appendix.

\subsection{Result Analysis}

\noindent\textbf{Q1.} \textit{Is our GIL framework and evaluation reasonable?} 

\begin{figure}[!t]
\centering\includegraphics[width=\linewidth]{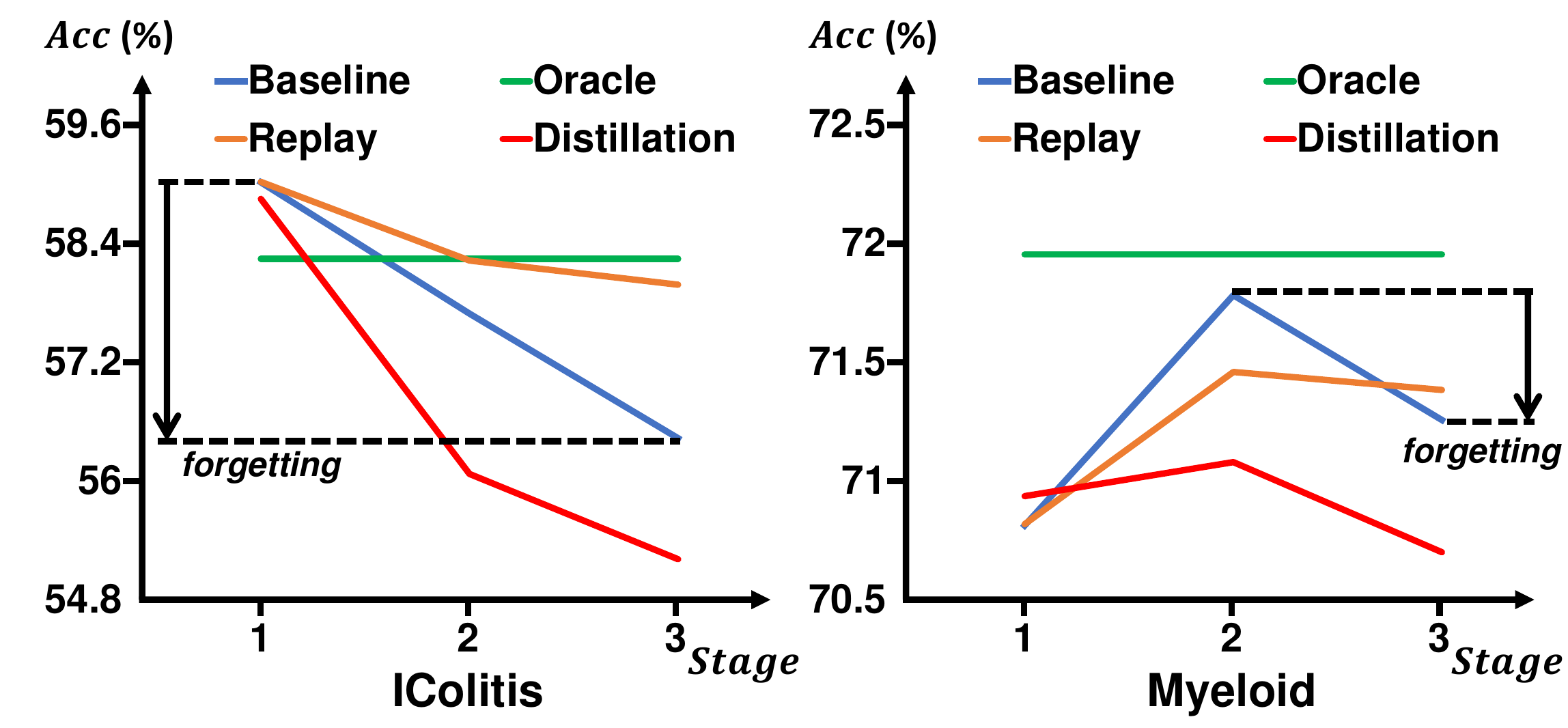}
    \caption{Test Accuracy (\%) for the 3-stage GIL setting (ICol-Myel-Panc) on the corresponding downstream classification datasets. The crucial genes for the last dataset Panc only learned in the last stage, thus there is no forgetting problem and we omit the results for Panc here.} 
    \label{fig:fig3stage}
\end{figure}

\noindent\textbf{A1.} The performance of the baseline in Table~\ref{table1}, Table~\ref{table3stage}, Table~\ref{table2}, and Figure~\ref{fig:fig3stage} demonstrate that, in the absence of any knowledge-preserving methods, the model progressively forgets previously learned genes. In the 2-stage GIL settings, the performance drops by 0.279 in regression and 1.816\% in downstream classification, on average, and drops by 0.181 in regression in the 3-stage GIL setting. This confirms the gene forgetting problem in GIL, justifying the effectiveness of our proposed GIL framework. Furthermore, by comparing Table~\ref{table1} and Table~\ref{table2}, we observe that both of the evaluations, gene-wise regression, and gene-based classification, show declines with the increase of stages when evaluating the baseline, indicating the consistency of our proposed evaluations.

\noindent\textbf{Q2.} \textit{Are the adapted methods we used effective?} 

\noindent\textbf{A2.} In Table~\ref{table1} and Table~\ref{table3stage}, we observe that both methods achieve improvements in gene-wise regression evaluation across different settings, indicating that they are effective and can prevent gene forgetting. As shown in Table~\ref{tableablation}, we also find that both methods consistently reduce gene forgetting, and as the hyperparameters increase, better performance is achieved, where the best performance is 0.018 and 0.154 for gene replay and gene distillation, respectively, under the selected scope of hyperparameters. 

Note that, for gene replay, as the number of replayed samples increases, it degenerates into the oracle method. For gene distillation, as $\lambda$ increases, the model may retain more previous knowledge, but it could impair the learning of new genes. For example, in Table~\ref{tableablation}, with $\lambda$ increasing from 0.5 to 10.0, the loss for Lupus increases from 0.134 to 0.143, demonstrating that learning for stage 2 is gradually failing.

{\scriptsize
\begin{table}[t]
\centering
\renewcommand{\arraystretch}{0.8}
{
\begin{tabular}{c|c|ccc}
\toprule
Method & Params & Norman & Lupus & $\Delta$ \\
\midrule
\multirow{2}{*}{Baseline} &  1 & 0.172 & - & - \\
&  2 & 0.424 & 0.134 & 0.253\\
\midrule
\multirow{4}{*}{Replay}  
&  $50$  & 0.293 & 0.136 & 0.121 \\
&  $10^2$  & 0.263 & 0.138 & 0.091 \\
&  $10^3$  & 0.215 & 0.134 & 0.043 \\
&  $10^4$   & 0.190 & 0.133 & 0.018 \\

\midrule
\multirow{4}{*}{Distill}   &  $0.5$ & 0.420 & 0.134 & 0.248 \\
&  $1.0$ & 0.402 & 0.135 & 0.230 \\
&  $5.0$ & 0.365 & 0.139 & 0.193 \\
&  $10.0$ & 0.326 & 0.143 & 0.154 \\
\bottomrule
\end{tabular}
}
\vspace{-4pt}
\caption{Averaged regression loss for ablation studies for gene replay and gene distillation under setting Norman-Lupus on the evaluation set. Except for the baseline method, all other methods only report the performance for stage 2. Params for baseline denote stages, for gene replay denotes the numbers of replayed samples, and for the distillation denote coefficients $\lambda$ in Eq.~\eqref{eq:distillation}. For the regression results, smaller values are better.}
\vspace{-6pt}
\label{tableablation}
\end{table}
}

\noindent\textbf{Q3.} \textit{Why does gene distillation perform well in gene-wise regression but poorly in gene-based classification?} 

\noindent\textbf{A3.} In Table~\ref{table1}, we find that the method gene distillation achieves an average $\Delta$ of 0.212 outperforming the baseline of 0.279. However, in Table~\ref{table2}, its average performance drop is 2.473\%, which is worse than the baseline of 1.816\%. This discrepancy indicates that gene distillation has limitations. A possible reason is that gene distillation indeed prevents gene forgetting on average, but degrades the features of the genes learned by the model. This conflict highlights the effectiveness and comprehensiveness of the two evaluations we proposed. Only methods that show consistent improvements across all evaluations could be considered robust.

\noindent\textbf{Q4.} \textit{Why are the oracle accuracies different for the same dataset across different settings?} 

\noindent\textbf{A4.} The reason is we removed crucial genes shared across downstream datasets when constructing different settings, to prevent cross-contamination of downstream classifications between datasets. For example, in Table~\ref{table1}, although both the Norman-Lupus and Lupus-Panc settings include the Lupus, the selected crucial genes for Lupus are different and thus induce the mentioned phenomenon.

\noindent\textbf{Q5.} \textit{Why have we only provided a few settings, and why is the number of stages limited?} 

\noindent\textbf{A5.} According to our proposed gene-based classification, we should identify genes that are crucial for each downstream dataset. However, all transcriptomic downstream datasets have a large amount of overlapping crucial genes. For example, in Table~\ref{table2}, the baseline achieves 67.313\% on Lupus at the first stage because some base genes and specific genes for Norman could help its classification. If a gene is crucial for many datasets but is assigned to a specific stage, the gene-based classification evaluation will fail. Thus, it is not as straightforward as just splitting two datasets into two stages to create a setting. In our experiment, we selected the above effective settings through preliminary experiments to serve as the benchmark settings for GIL, which is also one of our contributions.

\noindent\textbf{Q6.} \textit{What are the key differences between CIL and GIL according to the evaluations?} 

\noindent\textbf{A6.} The class as the label has a significant impact on classification performance, and new classes are exclusive from old classes, leading to the forgetting problem in CIL being easily observed and intuitively reflected in the results. However, for GIL, evaluating forgetting is more challenging. A gene is only a small part of the input, so measuring its significance is not obvious. For example, even if a gene is not learned, it may not influence the downstream tasks. Even though we construct the GIL framework and consistent evaluations, it is still difficult to observe a significant performance drop, as shown in Table~\ref{table2}, unlike CIL. Therefore, we propose the GIL framework, evaluations, and benchmark to illustrate the potential for this task and draw more attention to establishing better GIL evaluations and methods in this field.

\section{Conclusion}
\label{sec:conclusion}

In this paper, we introduce Gene Incremental Learning (GIL), a novel framework for single‑cell transcriptomics to address the problem of the gradual growth of genes. In this framework, we propose a series of novel designs to address the challenges in creating the GIL framework, including defining base genes to prevent the generation of semantically meaningless samples, and designing stage‑specific gene subsets alongside corresponding datasets, thereby establishing a semantic evaluation protocol under the transcriptomics settings. Through extensive experiments, we demonstrate the rationale behind our GIL framework and the effectiveness of the proposed evaluation protocol and method adaptations, establishing a comprehensive benchmark for GIL in single-cell transcriptomics. For future work, we will try to design specific GIL algorithms, and we also aim to extend the framework into other token-learning fields.

\section*{Acknowledgment}
This work was supported by the Strategic Priority Research Program of the Chinese Academy of Sciences under Grant No. XDA0460205.

\bibliography{reference}

\end{document}